\documentclass[final]{article}

\usepackage{neurips_2024}

\usepackage{natbib}
\usepackage[utf8]{inputenc} % allow utf-8 input
\usepackage[T1]{fontenc}    % use 8-bit T1 fonts
\usepackage{hyperref}       % hyperlinks
\usepackage{url}            % simple URL typesetting
\usepackage{booktabs}       % professional-quality tables
\usepackage{amsfonts}       % blackboard math symbols
\usepackage{nicefrac}       % compact symbols for 1/2, etc.
\usepackage{microtype}      % microtypography
\usepackage{xcolor}         % colors

\usepackage[english]{babel}
\usepackage{helvet}
\usepackage{etoolbox}
\usepackage{graphicx}
\usepackage{enumerate}
\usepackage{caption}
\usepackage{booktabs}
\usepackage{subcaption}
\usepackage{scrextend}
\usepackage{fancyhdr}
\newcounter{lemma}

\newcounter{theorem}

\usepackage{tabularx}
\usepackage{amsmath}
\usepackage{amsfonts}
\usepackage{enumitem}
\usepackage{multirow}
\usepackage{multicol}
\usepackage{enumitem}
\usepackage{float}
\usepackage{algpseudocode}
\usepackage{wrapfig}
\algrenewcommand\textproc{}

\title{Impact of Inaccurate Contamination Ratio on Robust Unsupervised Anomaly Detection}

\author{%
  Jordan F.~Masakuna\thanks{The first six authors are members of GRIC (\url{https://gric.recherche.usherbrooke.ca}), an interdisciplinary research group based at the University of Sherbrooke in Canada.} \\
  University of Sherbrooke\\
  Canada, QC J1K 2R1 \\
  University of Kinshasa\\
  Dem. Rep. of Congo\\
  \texttt{masj2413@usherbrooke.ca} \\
   \AND
   DJeff Kanda Nkashama \\
   University of Sherbrooke \\
   Canada, QC J1K 2R1 \\
   \texttt{nkad2101@usherbrooke.ca} \\
   \And
   Arian Soltani \\
   University of Sherbrooke \\
   Canada, QC J1K 2R1 \\
   \texttt{arian.soltani@usherbrooke.ca} \\
   \And
   Marc Frappier \\
   University of Sherbrooke \\
   Canada, QC J1K 2R1 \\
   \texttt{fram1801@usherbrooke.ca} \\
      \And
   Pierre-Martin Tardif \\
   University of Sherbrooke \\
   Canada, QC J1K 2R1 \\
   \texttt{tarp2202@usherbrooke.ca} \\
      \And
   Froduald Kabanza \\
   University of Sherbrooke \\
   Canada, QC J1K 2R1 \\
   \texttt{kabf2302@usherbrooke.ca} \\
}

\begin{document}

\maketitle

\begin{abstract}
Training data sets intended for unsupervised anomaly detection, typically presumed to be anomaly-free, often contain anomalies (or contamination), a challenge that significantly undermines model performance. Most robust unsupervised anomaly detection  models rely on contamination ratio information to tackle contamination. However, in reality, contamination ratio may be inaccurate. We investigate on the impact of inaccurate contamination ratio information in robust unsupervised anomaly detection. We  verify whether they are resilient to misinformed contamination ratios. Our investigation on 6 benchmark data sets reveals that such models are not adversely affected by exposure to misinformation. In fact, they can exhibit improved performance when provided with such inaccurate contamination ratios.

\end{abstract}

\section{Introduction and related work}
\label{sec:introduction}
Unsupervised anomaly detection relies heavily on the assumption that training data sets are devoid of anomalies. However, in practice, this assumption often falls short, as data sets frequently harbor anomalous instances \citep{jha2023label, perini2023estimating, qiu2022latent}, referred to as contamination \citep{hayes2018contamination,perini2023estimating}. The presence of contamination poses a significant challenge to unsupervised anomaly detection models, potentially compromising their effectiveness and reliability. To address this challenge, robust anomaly detection models have been developed (e.g., isolation forest (IF) \citep{liu2008isolation}, local outlier factor (LOF) \citep{alghushairy2020review}, one-class support vector machine (OCSVM) \citep{alam2020one}, neural transformation learning for deep anomaly detection beyond images (NeutrALAD) \citep{qiu2022latent} and deep unsupervised anomaly detection (DUAD) \citep{li2021deep}), aiming to mitigate the impact of contamination on model performance. Crucial to such models is the accurate estimation of contamination ratio, which informs the model about the proportion of anomalies within the data set. 

Contamination ratio is susceptible to inaccuracies, which can arise due to various factors such as data collection processes \citep{whang2023data, cowie2011issues} and labeling errors \citep{rottmann2023automated, northcutt2021confident}. The reliance of robust models on accurate contamination ratio information raises a critical question: How do these models perform when confronted with misinformed contamination ratios?
%
%In this study, we delve into the repercussions of inaccurate contamination ratio information on contamination-robust anomaly detection. 

 \begin{wrapfigure}{r}{0.4\textwidth}
 \centering
\includegraphics[width=5cm]{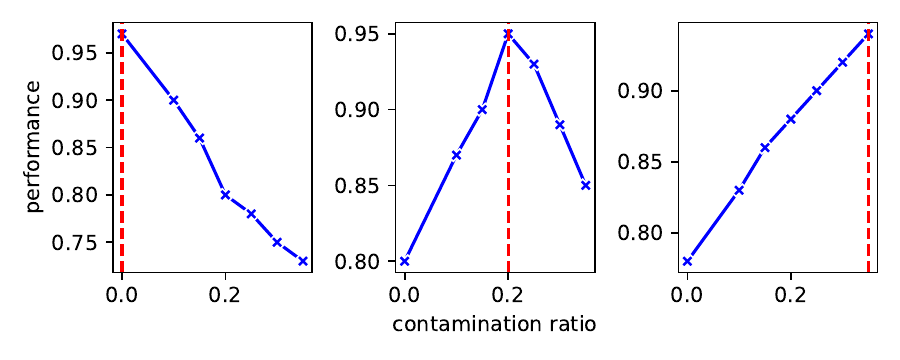}
\caption{Expected behavior. Red dashed line is true contamination ratio.}
\label{fig:expected}
\end{wrapfigure}
Our primary objective is to ascertain the resilience of contamination-robust models to misinformed contamination ratios. Through a meticulous investigation encompassing six benchmark data sets, we aim to shed light on the extent to which contamination-robust models are affected by inaccuracies in contamination ratio estimation. We expect model's performance to decline when received inaccurately specified contamination ratios (see Figure \ref{fig:expected}). 

This preliminary work focuses on shallow anomaly detection models only, i.e., IF, LOF and OCSVM.

%Through our empirical investigation and comparative analysis, we strive to contribute to the advancement of contamination-robust anomaly detection methodologies. By identifying the vulnerabilities and strengths of existing models in the context of misinformed contamination ratios, we aim to pave the way for the development of more resilient and reliable anomaly detection solutions capable of effectively handling real-world data complexities and uncertainties.

\section{Discussion and conclusion}
\begin{figure}[h]
\includegraphics[width=\textwidth, height=2cm]{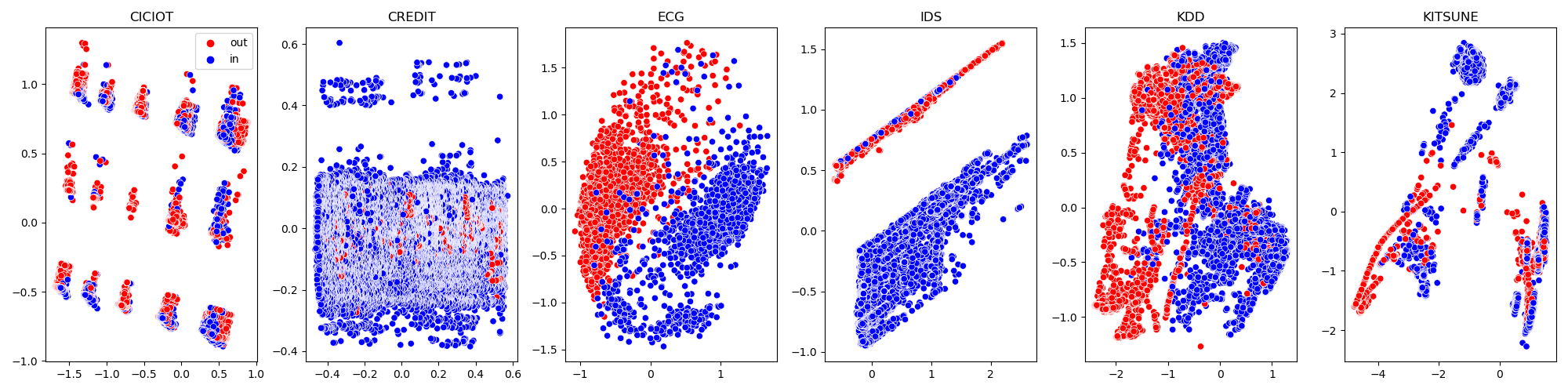}
\caption{Visualization of training data sets.}
\label{fig:training}
\end{figure}
We use six  data sets for experiments, whose the distributions are visualized in Figure \ref{fig:training}:
(1) \textbf{CICIOT} is a real-time data set containing 33  attacks that are executed in an IoT network \citep{neto2023ciciot2023}. 
(2) \textbf{CREDIT} is credit fraud data set comprising financial transaction records annotated with binary labels indicating fraudulent or legitimate transactions  \citep{warghade2020credit}. 
(3) \textbf{ECG} consists of recordings of electrical activity of the heart, capturing expected waveform patterns and anomalies \citep{khan2021ecg}. 
(4) \textbf{IDS}  contains simulated network traffics and several types of attacks \citep{sharafaldin2018toward}. 
(5) \textbf{KDD} contains simulated military traffics and several types of attacks \citep{tavallaee2009detailed}. 
(6) \textbf{KITSUNE}  is a network attack data set captured from either an IP-based commercial surveillance system or an IoT network \citep{mirsky2018kitsune}. 
For performance evaluation, we focus solely on accuracy since the test data subsets are well-balanced (in this binary classification task).

\begin{figure}[ht]
    \centering
        \includegraphics[width=\textwidth]{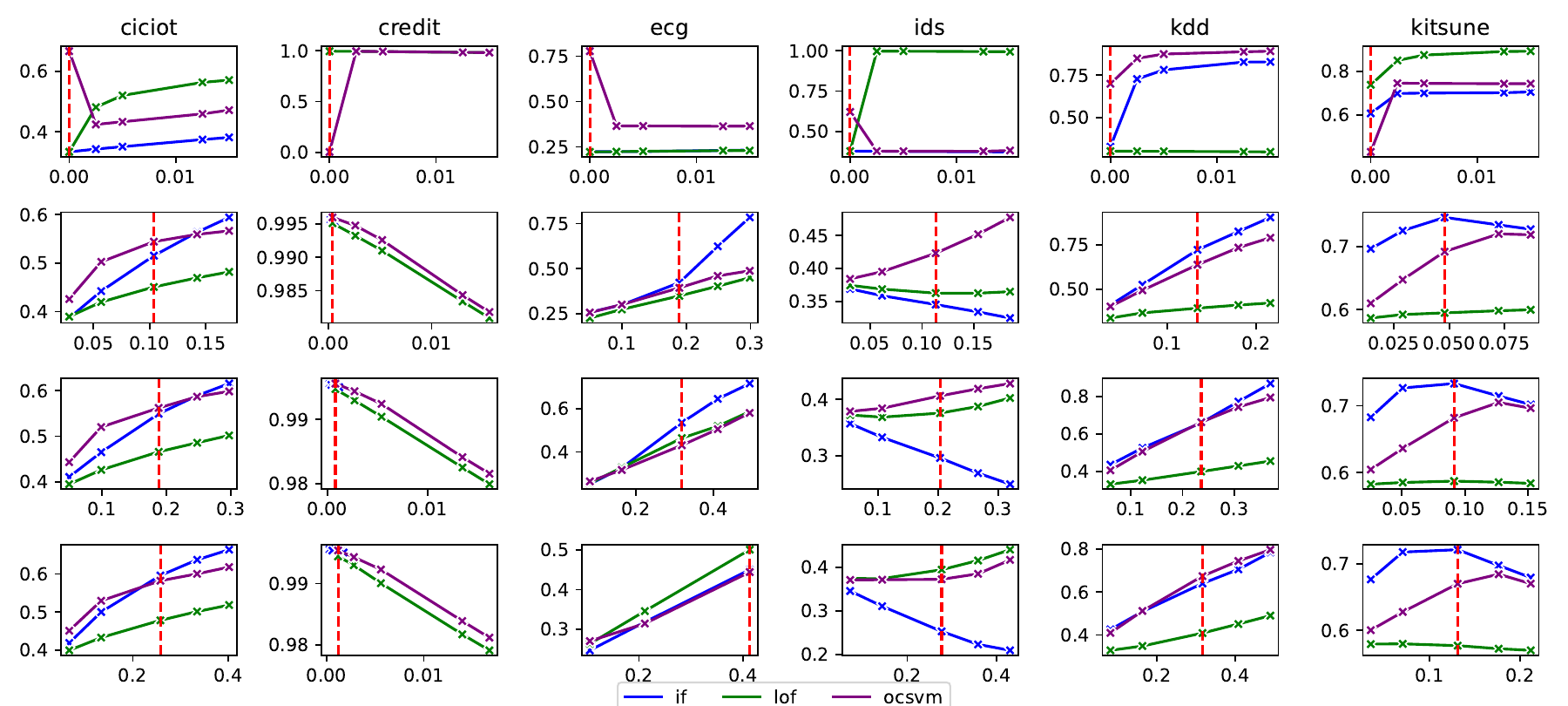}
    \caption{
    Model's accuracy against misinformed contamination ratio where x-axis represent misinformed contamination ratios. In red dashed  lines are true contamination ratios.}
    \label{fig:boxplot_bop_difference_all}
\end{figure}
Figure \ref{fig:boxplot_bop_difference_all} shows the accuracy of models against misinformed contamination ratios.
These findings  challenge the conventional assumption regarding the impact of misinformed contamination ratios on contamination-robust unsupervised anomaly detection models. Contrary to expectations, the tested models demonstrated resilience and even outperformed anticipated outcomes under such conditions.  These results suggest that the robust unsupervised anomaly detection models under consideration may not necessitate precise contamination ratios to address data contamination. This unexpected robustness suggests a need for deeper investigation into the underlying mechanisms of these models and highlights the importance of reassessing our assumptions in unsupervised anomaly detection research. Notably, while the expected behavior was observed on some instances (e.g., all 3 models on CREDIT and IF on IDS), the consistent trend across multiple data sets underscores the significance of these findings and their potential implications for advancing anomaly detection methodologies.

The implications of these unexpected findings are multifaceted and significant. They challenge the prevailing assumptions in anomaly detection research, prompting a reevaluation of the factors that influence model performance. This could lead to the refinement of existing methodologies and the development of more accurate anomaly detection systems. The implications of this study extend beyond anomaly detection alone, offering insights into broader issues of model robustness, reliability, and adaptability in the face of uncertainty in domains such as cybersecurity and fraud detection. Further investigation is needed to grasp the true meaning of these results, including deep models.

\bibliographystyle{plainnat}
\bibliography{output}

\end{document}